%% file: main.tex
\documentclass[letterpaper]{article} 
\usepackage{aaai2027}
\nocopyright
\usepackage[hyphens]{url}  
\usepackage{graphicx} 
\usepackage{natbib}  
\usepackage{caption} 
\usepackage{algorithm}
\usepackage{algorithmic}
\usepackage{amsmath}
\usepackage{newfloat}
\usepackage{listings}
\DeclareCaptionStyle{ruled}{labelfont=normalfont,labelsep=colon,strut=off} 
\floatstyle{ruled}
\newfloat{listing}{tb}{lst}{}
\floatname{listing}{Listing}

\usepackage{booktabs}
\usepackage[table]{xcolor}
\usepackage{multirow} 
\title{LD4WAM: Learning Latent Dynamics from Human Videos for World Action Models}
\author{
Zhenhao Shen\textsuperscript{\rm 1}\equalcontrib,
Jiaqi Liang\textsuperscript{\rm 1,2}\equalcontrib,
Jasper Lu\textsuperscript{\rm 1}\equalcontrib,
Feng Jiang\textsuperscript{\rm 1}\equalcontrib,
Yuran Wang\textsuperscript{\rm 1,2},
Chuanbo Wei\textsuperscript{\rm 1}, \\
Jiayi Liu\textsuperscript{\rm 1},
Jianchun Yang\textsuperscript{\rm 5},
Qize Yu\textsuperscript{\rm 1},
Jiadi You\textsuperscript{\rm 6},
Ce Hao\textsuperscript{\rm 4},
Guanqi He\textsuperscript{\rm 2},
Chen Xie\textsuperscript{\rm 3},
Ruihai Wu\textsuperscript{\rm 1}\corresponding
}
\affiliations{
{\large
\textsuperscript{\rm 1}Peking University \quad
\textsuperscript{\rm 2}WUJI \quad
\textsuperscript{\rm 3}Lightwheel \quad
\textsuperscript{\rm 4}Beijing Zhongguancun Academy \quad \\
\textsuperscript{\rm 5}Wuhan University \quad
\textsuperscript{\rm 6}The University of Hong Kong \\[0.6em]

{\Large Project page: \url{https://stubborn111.github.io/LD4WAM/}}
}

}

\begin{document}

\maketitle
\vspace*{-1.5em}

\begin{abstract}

Human video is playing an increasingly central role in training World Action Models (WAMs), owing to its diversity and low collection cost relative to teleoperated robot data. However, most WAMs learn from such video only by predicting pixel-level future frames, giving dynamics that are not directly actionable, whereas motion retargeting recovers directly actionable actions but leaves a large visual gap across embodiments.  We therefore propose motion-aligned latent dynamics as an embodiment-agnostic representation to bridge video priors and low-level actions. We further present LD4WAM, which pairs a Latent Dynamics Model trained with semantic reconstruction and real motion alignment with a World Dynamics Action Model built as a mixture-of-transformers (MoT), which preserves full future-video generation and uses learnable queries to distill these latent dynamics from generated futures for action conditioning. Pretrained on our curated unified dataset of over 5{,}000 hours of human and robot data, LD4WAM performs strongly in RoboTwin simulation and on real robots equipped with both grippers and dexterous hands, while generalizing well to unseen objects and backgrounds. 

\end{abstract}


\input{tex/Intro}

\input{tex/Related_Work}

\input{tex/Method}
\input{tex/Experiment}

\input{tex/Conclusion}

\bibliography{main}
\input{tex/Appendix}

\end{document}

%% file: tex/Intro.tex
\section{Introduction}

World Action Models (WAMs), which unify future observation prediction with action generation conditioned on it, have recently emerged as a leading paradigm for robot manipulation~\cite{li2026causalworldmodelingrobot, ye2026worldactionmodelszeroshot, yuan2026fastwam}. Since prediction can be supervised with video alone,  WAMs increasingly draw on large-scale egocentric human manipulation video, given its diversity and low collection cost relative to teleoperated robot demonstrations
~\cite{punamiya2026egoverseegocentrichumandataset, hoque2026egodexlearningdexterousmanipulation, grauman2022ego4dworld3000hours}. However, many existing WAMs~\cite{motubrainteam2026motubrainadvancedworldaction, bi2025motusunifiedlatentaction, beingbeyond2026beingh07} primarily use such video as pixel sequences supervised by future-frame reconstruction. As a result, the dynamics captured this way remain confined to the generative pathway and only weakly coupled to the policy, for lack of action labels to anchor them. This disconnect between prediction and action limits how effectively knowledge acquired from human video transfers to robot control.

A growing body of work therefore seeks to extract supervision from human video beyond pixel reconstruction. One line retargets human motion into the robot action space using hand or end-effector annotations from egocentric datasets~\cite{zheng2026egoscalescalingdexterousmanipulation, ma2026humanscaleegocentrichumanvideo}. However, such mappings can entangle transferable physical knowledge with embodiment-specific execution, introduce correspondence errors due to human-robot kinematic differences, and leave the visual embodiment gap unresolved. Another line learns latent actions from inter-frame pixel changes without action labels and uses them to condition robot control~\cite{ye2025latentactionpretrainingvideos, routray2025vipra, zhang2026clapcontrastivelatentaction}. Yet these codes are designed to explain pixel changes rather than isolate motion, so they may retain appearance- and embodiment-specific information while lacking a clear connection to executable motion. Human video therefore requires an intermediate representation that abstracts away appearance and embodiment differences, transfers across embodiments, and is grounded in real motion to support robot control.

\begin{figure*}[!t]
    \includegraphics[width=\textwidth]{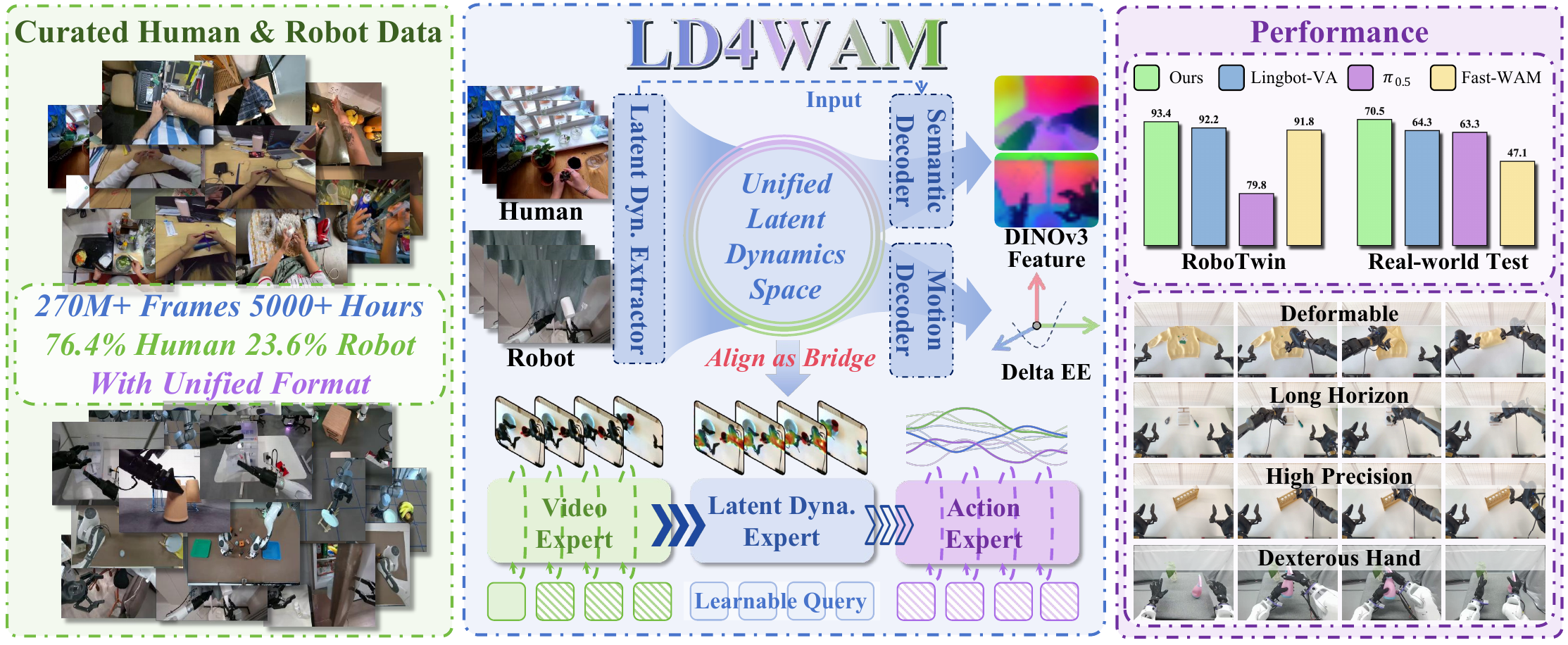}
    \caption{Overview of LD4WAM.
Left: a unified pretraining dataset of $5{,}000+$ hours human and robot videos. Middle: a Latent Dynamics Model distills
motion-aligned latent dynamics via semantic and motion (Delta EE) supervision; in the World Dynamics Action Model, learnable queries carry these latent
dynamics to bridge the video and action experts. Right: qualitative and quantitative results on RoboTwin and real robots demonstrate the strong manipulation capability of LD4WAM.}
    \label{fig:teaser}
\end{figure*}

We propose \emph{motion-aligned latent dynamics}, a compact representation of inter-frame dynamics learned in semantic feature space and explicitly aligned with observed motion. This representation provides an \textbf{embodiment-agnostic} bridge between high-dimensional video priors and low-level robot actions, enabling effective learning from both human and robot data. Semantic abstraction suppresses appearance- and embodiment-specific details, allowing the representation to capture dynamics shared across humans and robots. Motion alignment further grounds this representation in real end-effector motion, making it actionable for robot control.

Based on this, we present \textbf{LD4WAM}, which consists of two components (as shown in Fig.~\ref{fig:teaser}): 1. \textbf{Latent Dynamics Model (LDM)} first learns the bridge representation. In contrast to prior latent-action models that reconstruct pixels, the LDM is trained using next-frame \emph{semantic reconstruction} together with \emph{motion alignment}. These objectives encourage the latent space to capture motion-relevant semantic changes while suppressing appearance-specific variation. 2. \textbf{World Dynamics Action Model (WDAM)} then incorporates this representation while retaining full future-video generation. Learnable queries distill the LDM-defined latent dynamics from generated futures, and the resulting dynamics condition the action expert. The action expert therefore operates on a compact, motion-aligned dynamics signal rather than on raw pixels alone, while the model preserves the physical priors acquired by the pretrained video generator.

To enable pretraining at scale, we curate a corpus of human and robot manipulation data and process it through a unified cleaning pipeline. We filter out clips with excessive egocentric camera motion, hands outside the field of view, or insufficient manipulation motion. The remaining clips are standardized into a unified LeRobot format~\cite{lerobot}, with end-effector deltas expressed in a shared camera coordinate frame as motion labels. The dataset contains approximately \textbf{270M frames}, corresponding to more than \textbf{5{,}000 hours} of video. Pretrained on this corpus, LD4WAM performs strongly on the RoboTwin simulation benchmark and in real-robot experiments using both bimanual two-finger grippers and dexterous hands. It handles long-horizon, high-precision, and dexterous manipulation tasks while generalizing well to unseen objects and backgrounds.

Our main contributions are summarized as follows:
\begin{itemize}
    \item We introduce motion-aligned latent dynamics and a Latent Dynamics Model trained with semantic reconstruction and motion alignment, as an embodiment-agnostic bridge between video priors and low-level actions.
    \item We develop a World Dynamics Action Model that preserves full future-video generation and uses learnable queries to distill latent dynamics from generated futures for action conditioning.
    \item We curate a unified dataset comprising more than 5{,}000 hours of human and robot manipulation data and conduct extensive experiments in RoboTwin simulation and on real robots, demonstrating strong performance and generalization to unseen objects and backgrounds.
\end{itemize}

%% file: tex/Related_Work.tex
\section{Related Work}

\begin{figure}[t]
    \centering
    \includegraphics[width=\linewidth]{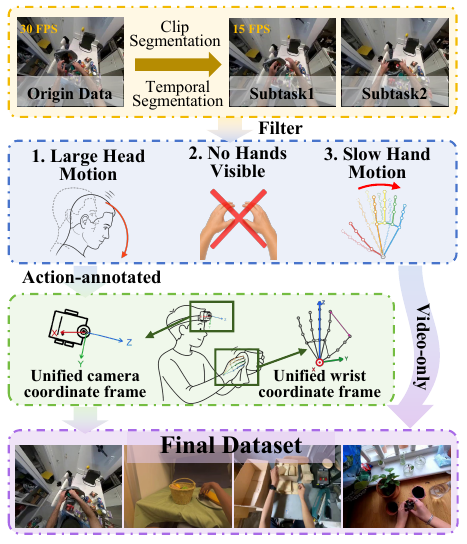}
    \caption{Human Data Process.
Long videos are segmented into subtask clips, downsampled to 15 FPS, and filtered.
Action-annotated clips are converted into unified camera and wrist coordinate
frames, while video-only clips go directly to the final dataset.}

    \label{fig:data}
\end{figure}

\subsection{Learning From Egocentric Human Videos }

Egocentric human video is a scalable source of manipulation behavior for
robot learning~\cite{punamiya2026egoverseegocentrichumandataset}, exploited along two lines.
The \emph{mapping} line retargets human demonstrations into a shared action
space and co-trains a policy on human and robot
data~\cite{zheng2026egoscalescalingdexterousmanipulation, ma2026humanscaleegocentrichumanvideo, yang2025egovlalearningvisionlanguageactionmodels, kareer2025emergencehumanrobottransfer, li2026aceego0unifyingegocentrichuman}; due to the embodiment gap it relies on
tight viewpoint, speed, and kinematics alignment, and breaks down for grippers
or dexterous hands. The \emph{learning} line instead uses human video as a pretraining corpus for transferable visual or dynamics
representations~\cite{kareer2024egomimicscalingimitationlearning, bi2025hrdthumanmanipulationenhanced, li2026egowamworldactionmodels, beingbeyond2026beingh05, routray2025vipra, zhang2026clapcontrastivelatentaction}, but the representation is never grounded in real motion, leaving the policy to recover the mapping from representation to action on its own from limited robot data. We follow the learning line, but ground the latent dynamics in real motion, which makes them directly usable for action.


\subsection{World Action Models for Robot Manipulation}

World action models predict the future observation and produce the action within a single model. Most of them do so in pixel space: they build on a pretrained video generator, which supplies physical priors learned from internet-scale video~\cite{ye2026worldactionmodelszeroshot, li2026causalworldmodelingrobot, yuan2026fastwam, bi2025motusunifiedlatentaction}. A second group moves the prediction target into richer feature spaces, using semantic feature~\cite{lyu2026lda, tang2026predictivealignedscalablerobot, beingbeyond2026beingh07, ma2026humanscaleegocentrichumanvideo} or 3D structure information~\cite{tian2026starryspatialtemporalactioncentricworld, guo2026unified4dworldaction, li2026wam4dfast4dworld}. However, these methods provide no representation that is both independent of embodiment and executable, limiting their use of unlabeled human video. LD4WAM learns such a representation in a semantic feature space, grounds it in real motion, and carries it to the action expert through learnable queries.



%% file: tex/Method.tex
\section{Method}

\begin{figure*}[t]
    \centering
    \includegraphics[width=\textwidth]{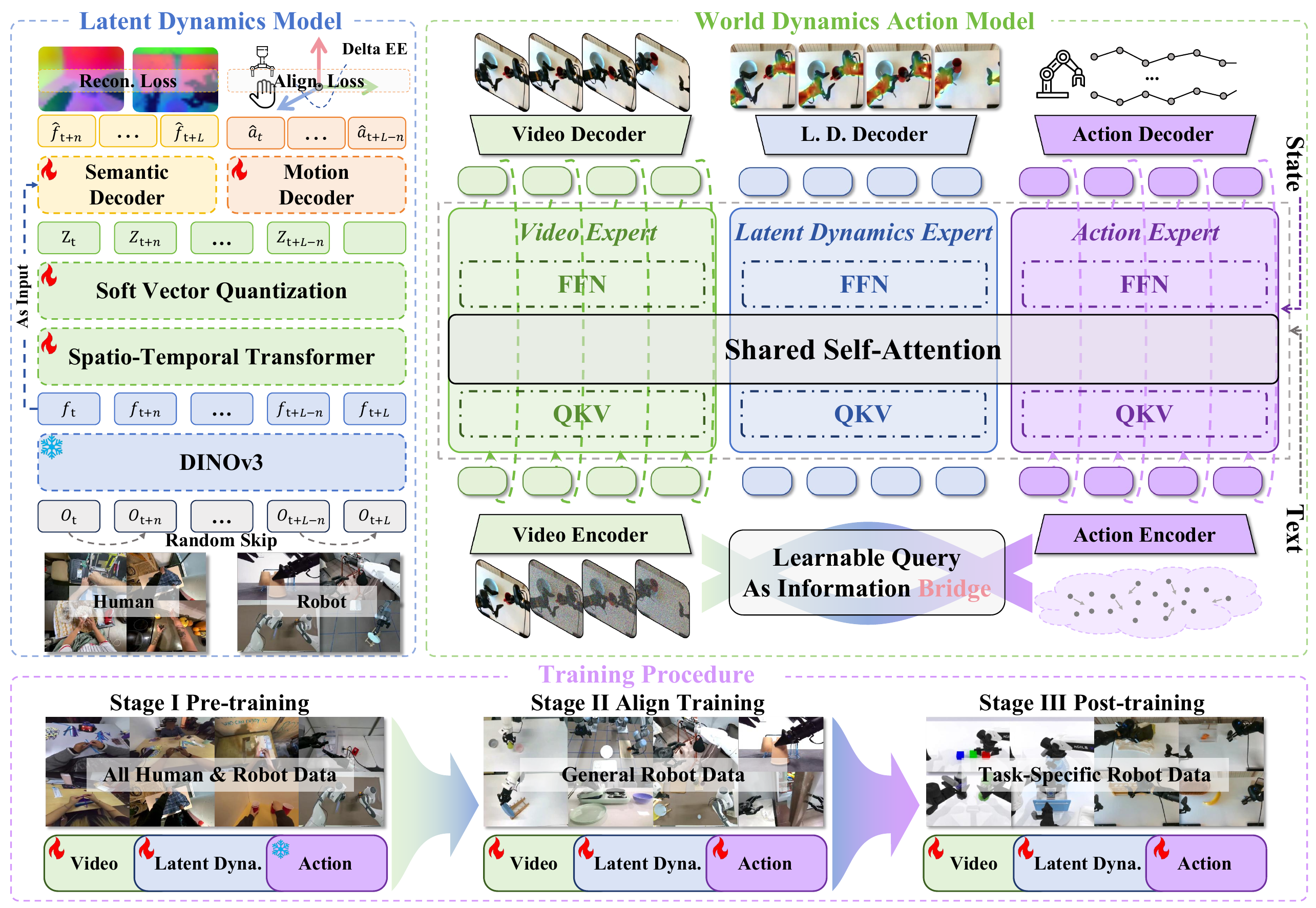}
    \caption{Pipeline of LD4WAM. Latent Dynamics Model (left): Frames sampled with random temporal skips are encoded by a frozen DINOv3; their features are aggregated by a spatio-temporal transformer and mapped via soft-VQ to latent dynamics $\{z_i\}$, supervised by semantic reconstruction and camera-frame end-effector deltas (Delta EE). World Dynamics Action Model (right): A Mixture-of-Transformers in which the video, latent dynamics, and action experts share a self-attention module. Learnable queries transfer the LDM-supervised latent dynamics from the video expert to the action expert; text conditions all three experts through cross-attention and is additionally concatenated with the state as input to the action expert. 
    Training Procedure (bottom): Stage~I pretrains the video and latent dynamics experts on all data while freezing the action expert; Stage~II aligns all experts on robot data; Stage~III post-trains the model on task-specific data.}
    \label{fig:pipeline}
    \vspace{-2mm}
\end{figure*}

As shown in Fig.~\ref{fig:pipeline}, LD4WAM comprises a \emph{Data Processing and Composition} pipeline and two models trained sequentially. Using the dataset curated by this pipeline, a \emph{Latent Dynamics Model} (LDM) learns a motion-aligned latent dynamics space that bridges visual observations and low-level actions. A \emph{World Dynamics Action Model} (WDAM) then predicts future observations while using these latent dynamics to ground action prediction. The complete \emph{Training Procedure} is described at the end of this section.

\subsection{Data Process and Composition}
\label{sec:data}
We process both human and robot datasets using the annotations provided by each dataset, and unify all sources into the LeRobot format. Long videos are segmented into subtask-level clips according to the annotated temporal boundaries. Clips shorter than 3~s or longer than 60~s are removed, and the remaining clips are uniformly downsampled to 15 FPS.

For the human datasets, we further apply the filtering pipeline in Fig.~\ref{fig:data}, handling two kinds of data. For \emph{action-annotated} data with head-camera extrinsics and
hand poses, we filter clips with large head motion (from the extrinsics), with both hands unannotated for an extended duration, or with little wrist motion over long windows; for each retained clip we express all actions in a unified head-camera coordinate frame and a unified wrist coordinate frame. For \emph{video-only} data without action labels, we apply optical-flow camera-shake filtering together with MediaPipe~\cite{lugaresi2019mediapipeframeworkbuildingperception} to keep only steady segments with visible, actively manipulating hands. For the robot datasets, we perform the corresponding format conversion into the same organization. All detailed procedures are provided in Appendix A.

Finally, we aggregate eight datasets spanning egocentric human videos and multi-embodiment robot demonstrations~\cite{contributors2024agibotworldrepo, robomind2, wu2025robomind1, egosuite, hoque2026egodexlearningdexterousmanipulation, punamiya2026egoverseegocentrichumandataset, grauman2022ego4dworld3000hours, xperience10m}, and unify them into
a single large-scale, cross-embodiment pretraining dataset. After the above process, the resulting dataset comprises 274.66M frames
(5{,}086 hours) at 15 FPS, of which human and robot data account for 76.4\% and 23.6\% of the frames, respectively. The detailed composition and per-dataset statistics are reported in Appendix B.


\subsection{Latent Dynamics Model}
\label{sec:ldm}
The LDM captures \emph{what changes} between frames while discarding static
appearance. Given a video, we sample a frame sequence $\{o_t, o_{t+n}, \dots,
o_{t+L}\}$ with a skip $n$ drawn from a set of temporal strides, which exposes the model to transitions at
varying temporal scales rather than a fixed frame rate. Each frame is encoded by a
frozen DINOv3~\cite{siméoni2025dinov3} into semantic features $f_i$; a spatio-temporal
transformer then aggregates them across space and time, and a \emph{soft vector
quantization}~\cite{SoftVQ} layer maps each transition into a latent-dynamics token $z_i$, whose
soft assignment to the codebook keeps the space smooth and structured while
suppressing low-level visual noise. Two heads supervise $z_i$ during training: a
\emph{semantic decoder} takes the current-frame feature $f_i$ together with $z_i$
and reconstructs the future features $\hat{f}_{i+n}$, forcing $z_i$ to encode the
semantic evolution from $f_i$ rather than appearance; a \emph{motion decoder}
predicts the inter-frame motion $\hat{m}_i$ as a delta end-effector pose, aligning
the dynamics with real motion. The objective is
\begin{equation}
\mathcal{L}_{\text{LDM}} = \mathcal{L}_{\text{sem}} + \lambda_m \mathcal{L}_{\text{mot}} + \mathcal{L}_{\text{vq}},
\end{equation}
\begin{equation}
\mathcal{L}_{\text{vq}} = \lambda_u \mathcal{L}_{\text{use}} + \lambda_e \mathcal{L}_{\text{ent}},
\end{equation}
where $\mathcal{L}_{\text{sem}}$ is a cosine distance in the frozen DINOv3 feature
space augmented by a small feature-norm matching term;
$\mathcal{L}_{\text{mot}}$ is applied only to frames that carry action labels,
so all video shapes the semantic dynamics while action-labeled data additionally
grounds them in executable motion;
and $\mathcal{L}_{\text{vq}}$ is a codebook regularizer comprising the KL divergence
from the mean code-assignment distribution to a uniform prior ($\mathcal{L}_{\text{use}}$)
and the mean per-token soft-assignment entropy ($\mathcal{L}_{\text{ent}}$),
jointly preventing codebook collapse.

At inference, we retain only the encoding path: given a frame sequence of the
training length, DINOv3, the spatio-temporal transformer, and the SoftVQ layer
produce the latent-dynamics tokens $\{z_i\}$, while both decoders are discarded.
These tokens are precomputed on the training videos and serve as the prediction
targets that supervise the WDAM's latent dynamics expert.

\subsection{World Dynamics Action Model}
\label{sec:wdam}
WDAM is a Mixture-of-Transformers architecture whose three experts keep separate weights and interact only through a shared self-attention over the concatenated token sequence. A \emph{video expert} and an
\emph{action expert} are flow-matching denoisers that generate the future video and
the future action, respectively; the video expert is initialized from a pretrained video
generation model Wan2.2~\cite{wan2025} to inherit its physical priors. Between them sits a \emph{latent
dynamic expert}: a clean stream of learnable queries, with no noise or diffusion
timestep, trained to reproduce the future latent dynamics under supervision from the
frozen LDM.

What turns these latent dynamics into an effective bridge is the asymmetric
visibility of the shared attention, which makes the layer stack a one-directional
pipeline $\text{video}\rightarrow\text{latent dynamics}\rightarrow\text{action}$.
The video stream attends only to itself, so future-video generation stays
self-contained and the pretrained prior is preserved unchanged, never perturbed by
the downstream streams. The learnable queries attend only to the predicted video and
themselves, distilling \emph{what will happen} into a compact summary while staying
blind to the action stream. The action stream then attends to both the predicted
video and this summary at every layer, so control is grounded on the purified latent
dynamics rather than left to read raw pixels alone. At inference this reduces to two
stages: the video is denoised first, and the action is then solved from the frozen
video together with the one-shot latent dynamics, so the bridge adds no extra
denoising steps.

For training, the three experts are supervised jointly,

\begin{equation}
\mathcal{L}_{\text{WDAM}} = \lambda_v \mathcal{L}_{\text{video}}^{\text{fm}}
+ \lambda_d \mathcal{L}_{\text{dyn}}^{\text{mse}}
+ \lambda_a \mathcal{L}_{\text{act}}^{\text{fm}},
\label{eq:wdam_loss}
\end{equation}
where $\mathcal{L}_{\text{video}}^{\text{fm}}$ and $\mathcal{L}_{\text{act}}^{\text{fm}}$
are flow-matching losses for the video and action denoisers, and
$\mathcal{L}_{\text{dyn}}^{\text{mse}}$ is a masked MSE between the predicted latent
dynamics and the offline LDM targets $\{z_i\}$. The loss weights $\lambda_v$, $\lambda_d$, and $\lambda_a$ are varied across the
three training stages to reflect their shifting objectives.

\subsection{Training Procedure}
\label{sec:train}
We train the LDM first on a high-quality subset obtained through stricter filtering and freeze it, then optimize the WDAM in three stages.
\textbf{Stage~I (Pre-training)} trains the video and latent dynamics experts on all human and robot data while freezing the action expert, which lacks labels on human video; this teaches the model to imagine the future and how to extract latent dynamics from it at scale. \textbf{Stage~II (Align Training)} unfreezes all experts and trains on general robot data, aligning the predicted latent dynamics with executable actions. \textbf{Stage~III (Post-training)} fine-tunes all experts on task-specific robot data for deployment.

Throughout, the DINOv3 encoder stays frozen and the video expert is initialized from a pretrained Wan2.2 generator. Full model dimensions, optimizer, learning rates, and per-stage schedules are provided in Appendix C.

%% file: tex/Experiment.tex
\section{Experiment}

\subsection{Setup}

\paragraph{Simulation.}
We evaluate on the RoboTwin benchmark~\cite{chen2025robotwin20scalabledata} extensively using its standard suite of 50 tasks. For each task we train on standard 50 clean and 500 randomized demonstrations, and train a single unified policy across all tasks. Each task is evaluated over 100 rollouts, and we report the success rate. We compare LD4WAM against state-of-the-art VLA and WAM methods, as well as baselines that leverage large-scale human data.

\begin{figure}[htbp]
    \centering
    \includegraphics[width=\linewidth]{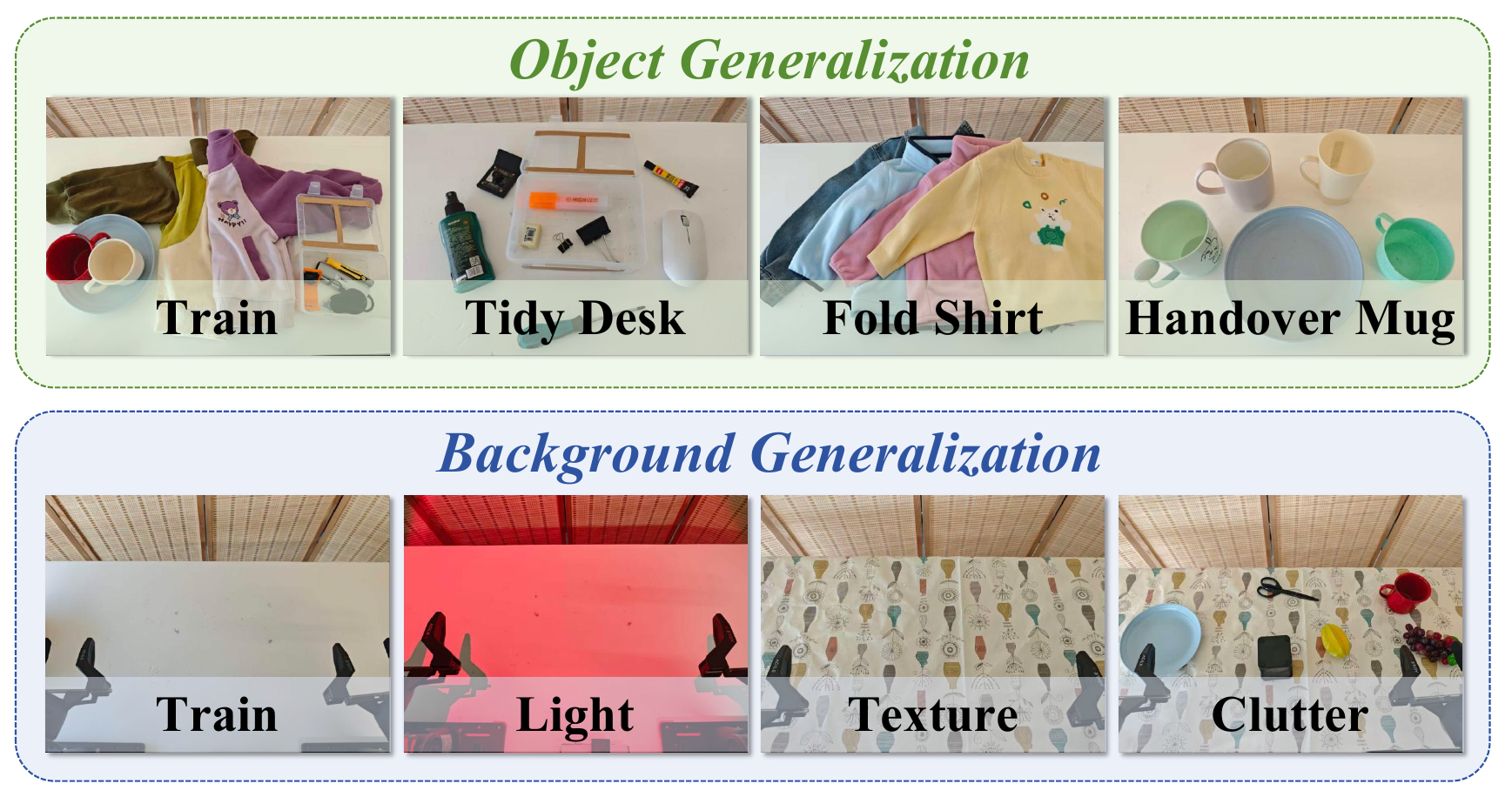}
    \caption{Generalization Setting.
We evaluate generalization along two axes: object and background.
Object Generalization (top) replaces the training objects with unseen ones for three tasks.
Background Generalization (bottom) perturbs the visual scene while keeping the task fixed, via lighting, table texture, and distractor objects.}
    \label{fig:gene_set}
    \vspace{-2mm}
\end{figure}

\paragraph{Real World.}
We deploy on two embodiments: a dual-arm PIPER equipped with parallel grippers, and
a dual-arm Tianji platform equipped with Wuji dexterous hands. On these we build
five gripper tasks and two dexterous-hand tasks that span a range of manipulation
skills: \emph{Sorting} (long-horizon pick-and-place by category), \emph{Shift Test
Tube} (high-precision insertion), \emph{Tidy Desk} (long-horizon multi-step
tidying), \emph{Fold Shirt} (deformable cloth manipulation), and \emph{Handover
Mug} (bimanual cross-arm coordination), together with two dexterous-hand tasks,
\emph{Place Rubik's Cube} (multi-finger grasp and placement)
and \emph{Spray Water} (in-hand tool use requiring independent finger actuation to
operate a trigger). To probe generalization, we further add \emph{object} and
\emph{background} generalization settings on Tidy Desk, Fold Shirt, and Handover
Mug, evaluating on unseen objects and unseen backgrounds; representative setups
for both settings are shown in Fig.~\ref{fig:gene_set}. We collect 50
demonstrations per task and evaluate each setting over 30 rollouts. Further task
details are in Appendix D. For baselines, we compare against
$\pi_{0.5}$~\cite{intelligence2025pi05visionlanguageactionmodelopenworld}, a state-of-the-art VLA, and Fast-WAM~\cite{yuan2026fastwam} and
Lingbot-VA~\cite{li2026causalworldmodelingrobot}, two leading world action models.

\begin{table}[htbp]
    \centering
    \caption{Performance comparison on RoboTwin.}
    \label{tab:sim}
    \resizebox{\columnwidth}{!}{
        \begin{tabular}{lccc}
            \toprule
            \textbf{Method} & \textbf{Clean} & \textbf{Random} & \textbf{Average} \\
            \midrule
            $\pi_{0.5}$~\cite{intelligence2025pi05visionlanguageactionmodelopenworld}      & 82.74           & 76.76            & 79.8             \\
            ACE-Ego-0~\cite{li2026aceego0unifyingegocentrichuman} & 91.12  & 90.62 &  90.9 \\
            Motus~\cite{bi2025motusunifiedlatentaction}      & 88.66             & 87.02            & 87.8             \\
            Lingbot-VA~\cite{li2026causalworldmodelingrobot}      & 92.90           & 91.50         & 92.2             \\
            Fast-WAM~\cite{yuan2026fastwam}      & 91.88          & 91.78            & 91.8             \\
            Being-H0.7~\cite{beingbeyond2026beingh07}      & 90.20          & 89.60            & 89.9             \\
            LaWAM~\cite{chen2026lawamlatentworldaction}      & 92.64           & 89.80          & 91.2            \\
            Ours        & \textbf{93.96}  & \textbf{92.78}   & \textbf{93.4}    \\
            \bottomrule
        \end{tabular}
    }
    \vspace{-2mm}
\end{table}

\subsubsection{Latent Dynamics Regression.}
Because the latent dynamics cannot be scored directly by task success, we measure
how much action-relevant information they carry by regressing the end-effector
motion from them. We freeze the LDM, extract latent dynamics from video clips, and
train a shallow head to read out the per-step end-effector delta; being shallow, it
probes what the latent already encodes rather than learning new features, and we
report the scale-normalized MSE. We build the evaluation set from RoboTwin over
all $50$ tasks, using $2{,}500$ clean and $2{,}500$ randomized episodes, cut into non-overlapping
$8$-frame clips split $8{:}2$ into probe-training and validation. To test temporal robustness, we sample the clips at several
frame strides and group them into a \emph{high-rate} band ($12.5\text{--}50$\,Hz)
and a \emph{low-rate} band ($2.5\text{--}10$\,Hz), which probe regression over small
and large motions, respectively. For each stride we train a separate head under the
same configuration; further details are in Appendix E.

\begin{figure*}[t]
    \centering
    \includegraphics[width=\textwidth]{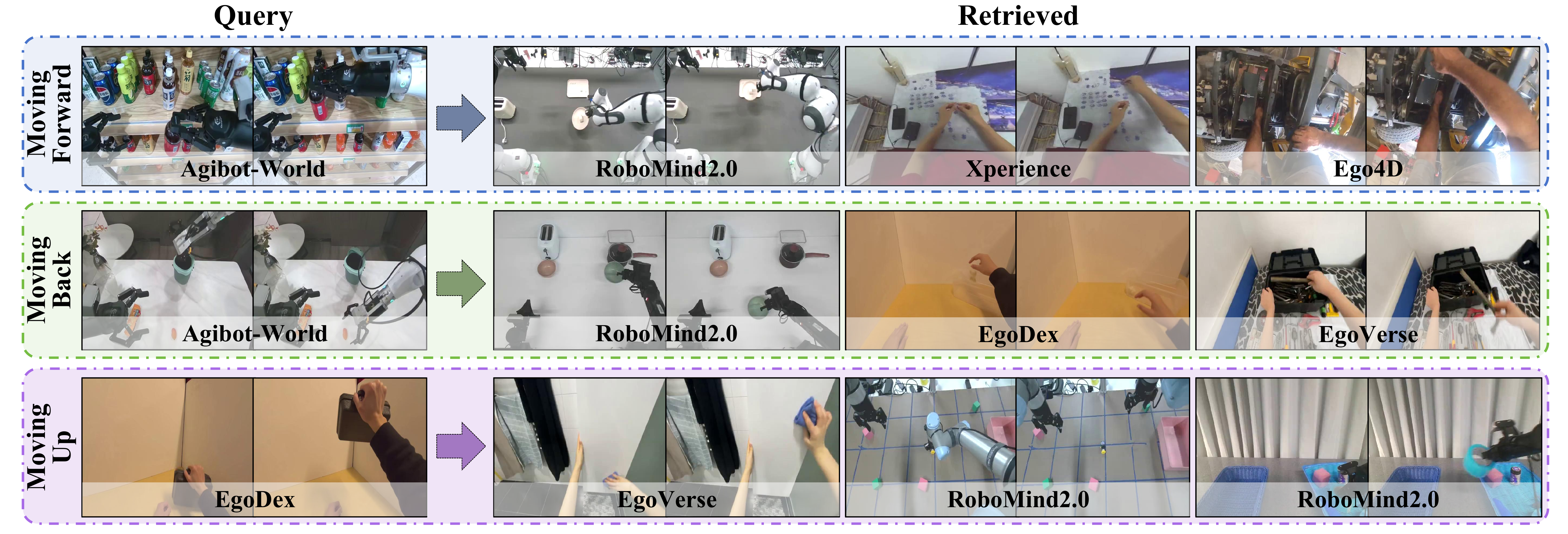}
    \caption{Cross-domain latent dynamics retrieval. Nearest neighbors of each query clip (left) by cosine similarity in the frozen LDM latent space, pooled across human and robot data; the source dataset is labeled under each panel. }
    \label{fig:retrieval}
\end{figure*}

\begin{table*}[htbp]
\centering
\caption{Success rate (\%) on real-world gripper and dexterous-hand manipulation tasks.}
\label{tab:main_results}
\setlength{\tabcolsep}{5pt}
\resizebox{\textwidth}{!}{%
\begin{tabular}{@{}l ccccc cc c@{}}
\toprule
& \multicolumn{5}{c}{Gripper} & \multicolumn{2}{c}{Dexterous Hand} & \\
\cmidrule(lr){2-6}\cmidrule(lr){7-8}
Method
 & Sorting & Shift Test Tube & Tidy Desk & Fold Shirt & Handover Mug
 & Place Rubik's Cube & Spray Water & Average \\
\midrule
$\pi_{0.5}$ & 63.3 & 43.3 & \textbf{73.3} & 76.7 & 70.0 & 76.7  & 40.0 & 63.3 \\
Fast-WAM     & 50.0 & 10.0 & 53.3 & 46.7 & 63.3 & 76.7 & 30.0 & 47.1 \\
Lingbot-VA   & \textbf{80.0} & 40.0 & 63.3 & 73.3 & 76.7 & 80.0 & 36.7 & 64.3 \\
\textbf{Ours} & \textbf{80.0} & \textbf{50.0} & 70.0 & \textbf{80.0} & \textbf{83.3} & \textbf{83.3} & \textbf{46.7} & \textbf{70.5} \\
\bottomrule
\end{tabular}%
}
\end{table*}

\subsection{Result Analysis}

\subsubsection{Latent Dynamics Retrieval.}
As shown in Fig.~\ref{fig:retrieval}, we retrieve the nearest neighbors of a two-frame
query by cosine similarity of latent dynamics across different human and robot datasets; the query is a robot
arm under opposite motions in rows 1 and 2 and a human hand in row 3. The retrieved
neighbors consistently reproduce the query's motion, and the two opposite queries retrieve
neighbors that move in correspondingly different ways. This pattern holds not only across robot embodiments but even between a human hand and a
robot arm sharing the consistent latent dynamics.

\subsubsection{Simulation.}
As shown in Tab.~\ref{tab:sim}, LD4WAM reaches an average success rate of 93.4\%
across the 50 tasks on the RoboTwin benchmark, outperforming the strong VLA
baseline ACE-Ego-0, which is also trained on human data, by $+2.5$ points and the leading WAM Lingbot-VA by $+1.2$ points. The top methods are tightly clustered in simulation, so we treat
this benchmark as a competitiveness check and defer the discriminative comparison
to the real world.

\subsubsection{Real World.} 
\begin{figure*}[t]
    \centering
    \includegraphics[width=\textwidth]{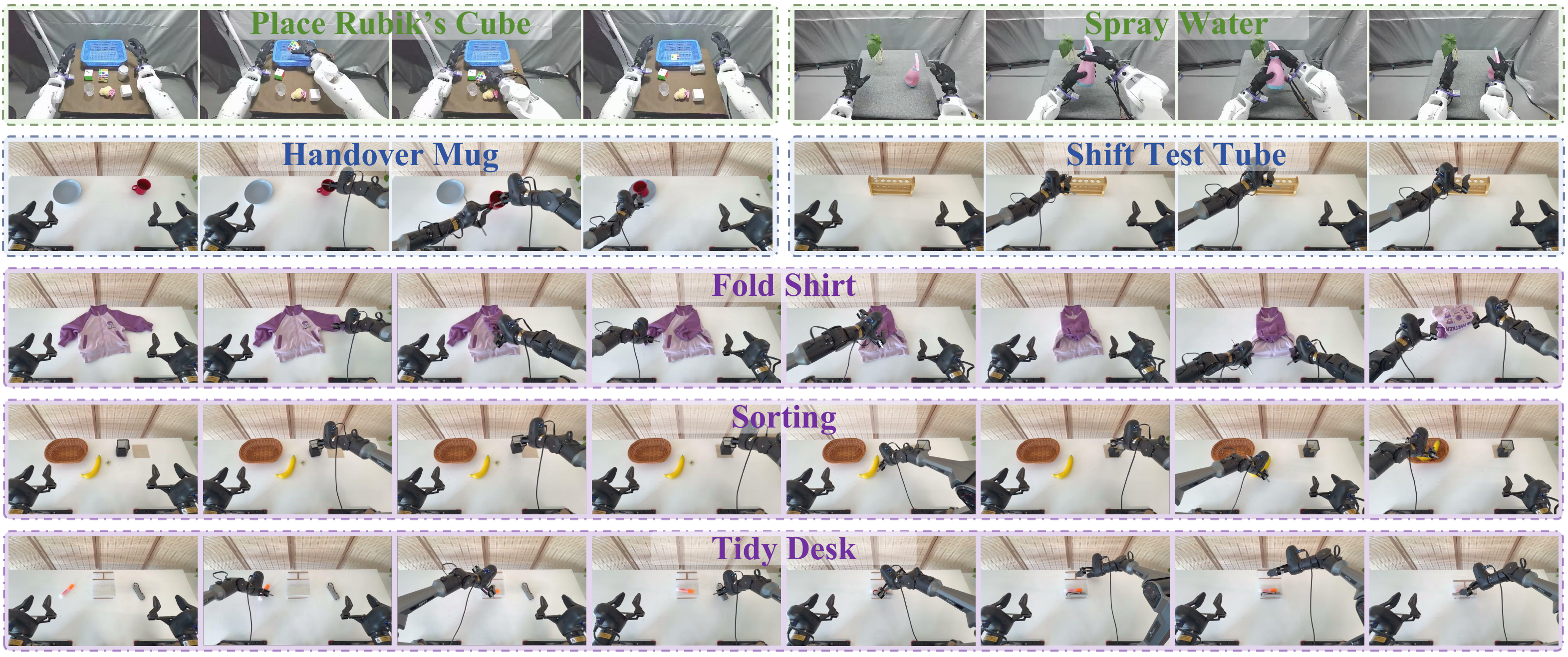}
    \caption{Real-world Execution. Qualitative real-world rollouts of our policy across all evaluation tasks,
spanning both dexterous-hand and gripper manipulation.}
    \label{fig:realdeployment}
    \vspace{-1mm}
\end{figure*}

As shown in Tab.~\ref{tab:main_results}, LD4WAM attains the best
success rate on six of the seven real-world tasks across both embodiments, raising the average from 63.3\%, 47.1\%, and 64.3\% for $\pi_{0.5}$, Fast-WAM, and Lingbot-VA to 70.5\%. Fig.~\ref{fig:realdeployment} further provides qualitative rollouts on each task.

\subsubsection{Generalization.}
Beyond the in-distribution comparison, Tab.~\ref{tab:ablation} evaluates both
generalization settings. On unseen objects, LD4WAM retains 88.6\% of its
in-distribution performance, the highest of all methods, confirming that
motion-aligned latent dynamics are largely invariant to \emph{which} object is
manipulated. Background shift is more revealing: although LD4WAM improves over
prior world action models by $+10.0$ and $+34.4$ points, it still trails
$\pi_{0.5}$ (44.4\% vs.\ 54.5\%), with the gap concentrated on texture-level
perturbations. 
We attribute this to a structural property of the world-action paradigm: control is
conditioned on generated future video, and an unseen texture is precisely what video prediction handles worst, so the action expert must act on a degraded rollout, whereas a VLA inherits texture-invariant features from an internet-scale backbone. The latent dynamics mitigate this by discarding much of the appearance
variation in a frozen semantic feature space, but they are read out from the same generated video, so the effect is suppressed rather than removed. Appendix G provides additional qualitative rollouts under object and background shifts.

\begin{table}[htbp]
\centering
\caption{Ablation of LDM under latent dynamics regression (scale-normalized MSE $\downarrow$, averaged within each band).}
\label{tab:lam}
\resizebox{\columnwidth}{!}{%
\begin{tabular}{@{}lccccc@{}}
\toprule
Rate band & \textbf{LDM} & w/o motion align. & w/o multi-stride & w/ pixel recon. & w/ hard VQ \\
\midrule
High-rate  & \textbf{0.21} & 0.78  & 0.24 & 0.25 & 0.37 \\
Low-rate  & \textbf{2.13} & 10.59 & 2.54 & 2.69 & 4.00 \\
\bottomrule
\end{tabular}%
}
\vspace{-2mm}
\end{table}

\begin{table*}[t]
\centering
\renewcommand{\arraystretch}{0.8}
\setlength{\tabcolsep}{8pt}
\caption{Generalization and architecture ablations (success rate, \%). ``Obj''/``Bg'' denote object/background generalization. Bold marks the best result per column. ``\texttt{+}'' cumulatively adds a component that is retained thereafter; ``$\hookrightarrow$'' replaces the component added immediately above with an alternative that is neither stacked nor carried forward. The last row is the full model.}
\label{tab:ablation}
\begin{tabular}{@{}l ccc ccc ccc c@{}}
\toprule
\multirow{2}{*}{Method}
 & \multicolumn{3}{c}{Tidy Desk} & \multicolumn{3}{c}{Fold Shirt} & \multicolumn{3}{c}{Handover Mug}
 & \multirow{2}{*}{Average} \\
\cmidrule(lr){2-4}\cmidrule(lr){5-7}\cmidrule(lr){8-10}
 & Base & Obj & Bg & Base & Obj & Bg & Base & Obj & Bg & \\
\midrule
$\pi_{0.5}$ & \textbf{73.3} & 53.3 & \textbf{56.7} & 76.7 & 60.0 & \textbf{50.0} & 70.0 & 66.7 & \textbf{56.7} & 62.6 \\
Fast-WAM     & 53.3 & 30.0 & 13.3 & 46.7 & 23.3 & 6.67 & 63.3 & 33.3 & 10.0 & 31.1 \\
Lingbot-VA   & 63.3 & 43.3 & 30.0 & 73.3 & 63.3 & 33.3 & 76.7 & 70.0 & 40.0 & 54.8 \\
\midrule
\multicolumn{11}{@{}l}{\textit{Ablation of our architecture}} \\
Video-Action Dual-Expert WAM
    & 56.7 & 30.0 & 13.3 & 63.3 & 56.7 & 10.0 & 66.7 & 50.0 & 13.3 & 40.0 \\
+ Latent Dynamic Expert (Ours)
    & 63.3 & 40.0 & 20.0 & 73.3 & 63.3 & 16.7 & 73.3 & 63.3 & 20.0 & 48.1 \\
\quad$\hookrightarrow$ LDM w/o motion alignment
    & 56.7 & 36.7 & 16.7 & 70.0 & 60.0 & 13.3 & 66.7 & 56.7 & 13.3 & 43.3 \\
\quad$\hookrightarrow$ LDM w/ pixel reconstruction
    & 60.0 & 36.7 & 13.3 & 66.7 & 63.3 & 16.7 & 66.7 & 60.0 & 16.7 & 44.5 \\
+ Pre-training
    & 63.3 & 53.3 & 40.0 & 76.7 & 66.7 & 40.0 & 73.3 & 70.0 & 46.7 & 58.9  \\
\textbf{+ Align Training (Full)}
    & 70.0 & \textbf{56.7} & 43.3 & \textbf{80.0} & \textbf{70.0} & 40.0 & \textbf{83.3} & \textbf{80.0} & 50.0 & \textbf{63.7} \\
\bottomrule
\end{tabular}
\vspace{-1mm}
\end{table*}

\subsection{Ablation Analysis}

\subsubsection{Latent Dynamics Model Ablation.}
Tab.~\ref{tab:lam} ablates the four LDM ingredients under the latent dynamics regression.
Motion alignment is the key one, tying the latent dynamics to real end-effector motion so
they track body motion rather than background or appearance change; removing it inflates the
error by $3.7\text{--}5.0\times$, by far the largest effect. Soft VQ keeps the space
continuous to match continuous motion, and hard quantization still adds $76\text{--}88\%$.
Semantic reconstruction keeps the latent dynamics on high-level content, whereas
reconstructing pixels ties them to noisy low-level appearance. Multi-stride sampling covers
both large and small inter-frame motions, whereas a single scale
covers only a narrow range.


\subsubsection{Full Architecture Ablation.}
Tab.~\ref{tab:ablation} ablates the full architecture by real-world success rate, building up from a Video-Action Dual-Expert WAM baseline (40.0\%). Adding the Latent Dynamic
Expert raises the average success rate to 48.1\%. Its benefit hinges on how the LDM targets are supervised, and ablating either of the two signals that shape them degrades control: removing motion alignment, so the latent dynamics are never grounded in executable motion, drops the average to 43.3\%, while replacing
the semantic feature-space objective with pixel reconstruction drops it to 44.5\%. Pretraining adds a further 10.8 points, and is almost entirely a generalization gain: the in-distribution average moves by only 1.1 points, while unseen objects improve by 7.8 points and unseen backgrounds by 23.3 points, indicating that
it supplies the visual and dynamic coverage the task-specific robot data lacks.
Finally, the alignment middle-training stage yields the full model at 63.7\%, adding 4.8 points overall; here the gains concentrate on execution quality and object generalization, showing that the two stages are complementary, alignment sharpens how the latent dynamics are consumed by the action expert, pretraining broadens what those dynamics have seen. Together they lift the baseline by 23.7 points.

%% file: tex/Conclusion.tex

\section{Conclusion}
We presented LD4WAM, a world action model that turns large-scale human and
robot video into transferable, generalizable manipulation capabilities through
motion-aligned latent dynamics. We first curate a unified dataset of over 5{,}000
hours of human and robot data into a shared format. On this dataset, a Latent
Dynamics Model distills this representation via semantic reconstruction and
real-motion alignment, and a World Dynamics Action Model uses learnable queries
to read it out as a signal that conditions action. Across RoboTwin and two
real-robot embodiments, LD4WAM outperforms strong VLA and world action model
baselines, with strong generalization to unseen objects and backgrounds.

%% file: tex/Appendix.tex
\clearpage
\section*{Appendix}

\subsection{A. Data Processing Details}
\label{app:dataprocess}

\paragraph{Segmentation and temporal filtering.}
All source videos are segmented into subtask-level clips using the temporal boundaries
provided by each dataset's annotations. We discard clips shorter than 3~s or longer
than 60~s, and uniformly downsample the remainder from their native frame rate to
15~FPS.

\paragraph{Unified coordinate frames.}
For action-annotated human data, all actions of a clip are expressed relative to the
head camera at the first frame, giving a \emph{unified camera coordinate frame} that
removes absolute pose and makes trajectories comparable across clips and datasets. In parallel, each hand is mapped to an end-effector in a \emph{unified wrist coordinate frame}. Let $p_w$, $p_m$, $p_i$, and $p_l$ be the keypoints of the wrist, the
middle-finger knuckle, the index knuckle, and the little-finger knuckle, and let
$\widehat{v} = v / \lVert v \rVert$. The frame is defined as follows:
\begin{itemize}
    \item The origin is placed at $p_w$.
    \item $z = \widehat{p_m - p_w}$, pointing from the wrist along the hand.
    \item $x = \widehat{z \times (p_l - p_i)}$, the palmar normal, with the sign of
    $(p_l - p_i)$ taken per hand so that $x$ always points from the back of the hand
    toward the palm.
    \item $y = z \times x$, along the knuckle line.
\end{itemize}

For the robot data, the same construction is applied to the gripper. Let $q_b$ be the
point at which the gripper attaches to the arm, $q_t$ the midpoint between the
fingertips, and $q_1$, $q_2$ the two jaw positions. The frame is defined as follows:
\begin{itemize}
    \item The origin is placed at $q_b$.
    \item $z = \widehat{q_t - q_b}$, pointing from the base along the gripper.
    \item $x = \widehat{z \times (q_2 - q_1)}$, normal to the plane in which the jaws
    travel, with the labelling of $q_1$ and $q_2$ fixed per embodiment so that $x$
    agrees with the end-effector frame declared by the robot and is the same for both
    arms.
    \item $y = z \times x$, along the jaw opening direction.
\end{itemize}
The two constructions assign the same physical meaning to each axis. The exported motiond is the frame-to-frame difference of this pose.

\paragraph{Filtering criteria.}
For action-annotated data, we remove a clip if (i) the head-camera extrinsics indicate
large head motion, (ii) both hands lack valid annotations for an extended duration, or
(iii) the wrist moves little over a long temporal window. 
For video-only data, which has no extrinsics or action labels, we run MediaPipe hand detection and keep only segments in which at least 50\% of frames contain a detected hand and the wrist trajectory spans more than 8 px. We additionally compute sparse Lucas–Kanade optical flow (Shi–Tomasi corners, 3-level pyramid) as a proxy for camera motion, and discard them if the mean per-frame flow magnitude over the clip exceeds 20 px (at the 15 fps sampling rate) or its range exceeds 80 px.

\subsection{B. Dataset Details}
Tab.~\ref{tab:ego_dataset_v2} lists the pre-training datasets: five egocentric
human and three multi-embodiment robot datasets, unified into the LeRobot format
at 15 FPS, with human data forming the majority of the corpus.
\label{app:datainfo}
\begin{table}[t]
\centering
\caption{Pre-training Dataset Statistics (FPS = 15).}
\label{tab:ego_dataset_v2}
\resizebox{\columnwidth}{!}{%
\begin{tabular}{@{}llrr@{}}
\toprule
Category & Dataset & Frames\,(M) & Hours \\
\midrule
\multirow{4}{*}{Human}
 & Ego4D~\cite{grauman2022ego4dworld3000hours}      & 99.88 & 1849.7 \\
 & EgoVerse~\cite{punamiya2026egoverseegocentrichumandataset}   & 50.96 &  943.6 \\
 & EgoDex~\cite{hoque2026egodexlearningdexterousmanipulation}     & 35.67 &  660.6 \\
  & EgoSuite~\cite{egosuite}  &  16.45 &  304.7 \\
 & Xperience~\cite{xperience10m}  &  6.93 &  128.2 \\
\cmidrule(l){2-4}
 & \textbf{Subtotal} & \textbf{209.89} & \textbf{3886.9} \\
\midrule
\multirow{5}{*}{Robot}
 & Agibot-World~\cite{contributors2024agibotworldrepo}    & 30.23 & 559.8 \\
 & RM2.0-Franka~\cite{robomind2}    & 21.00 & 388.9 \\
 & RM2.0-Agilex~\cite{robomind2}  &  6.92 & 128.2 \\
 & RM2.0-UR~\cite{robomind2}      &  3.88 &  71.8 \\
 & RM1.0-Franka~\cite{wu2025robomind1}  &  2.73 &  50.6 \\
\cmidrule(l){2-4}
 & \textbf{Subtotal} & \textbf{64.77} & \textbf{1199.4} \\
\midrule
\multicolumn{2}{@{}l}{\textbf{Total}} & \textbf{274.66} & \textbf{5086.3} \\
\multicolumn{2}{@{}l}{Frame ratio} & \multicolumn{2}{r}{H 76.4\% / R 23.6\%} \\
\bottomrule
\end{tabular}%
}
\end{table}

\subsection{C. Training Details}
\paragraph{Latent Dynamics Model.}
The spatial encoder is a frozen DINOv3 ViT-L/16, producing 196 patch tokens of dimension 1024 per frame. The spatio-temporal transformer has 6 non-causal layers (encoder) and 6 causal layers (decoder), each with 16 attention heads and head dimension 64. The SoftVQ codebook contains 8 entries; each transition is quantized into a $4{\times}4$ grid of 16 code slots, each of dimension 32, yielding a 512-dimensional latent vector per transition. 

We find that large motion noise degrades the LDM training, so we tighten the
thresholds of our data filtering pipeline (App.~A\ref{app:dataprocess}) and re-filter the corpus into a cleaner
subset of roughly $1{,}500$ hours to train the LDM for semantic reconstruction. Motion
alignment is applied to the four datasets with verified per-step delta end-effector
labels (EgoVerse, EgoDex, Xperience, and Agibot-World).

Motion alignment is introduced after a warmup phase to allow semantic dynamics to stabilize before grounding. The model is trained with AdamW ($\text{lr}{=}3{\times}10^{-5}$, cosine decay, weight decay $0.01$, gradient clip $1.0$).

\begin{table}[t]
  \centering
  \small
  \resizebox{\columnwidth}{!}{%
  \begin{tabular}{ll}
    \toprule
    \textbf{Hyperparameter} & \textbf{Value} \\
    \midrule
    Optimizer            & AdamW $(\beta_1,\beta_2){=}(0.9,0.95)$ \\
    Weight decay         & $0.01$ \\
    Gradient clip        & $1.0$ (global norm) \\
    LR schedule          & cosine, $5\%$ warmup \\
    Peak learning rate   & $1\times10^{-4}$ \\
    Min learning rate    & $1\%$ of peak \\
    Precision            & \texttt{bf16} \\
    Sharding             & DeepSpeed ZeRO-2 \\
    Grad.\ checkpointing & enabled \\
    Seed                 & $42$ \\
    Frozen modules       & T5 text encoder, VAE, LDM \\
    \midrule
    \multicolumn{2}{l}{\textit{Stage~I (Pre-training)}} \\
    \quad Trainable experts & video, latent dynamics \\
    \quad Views             & $1$ (head) \\
    \quad Batch size / GPU  & $128$ \\
    \quad Schedule          & $1$ epoch \\
    \quad $(\lambda_v,\lambda_d,\lambda_a)$ & $(1.0,\,1.0,\,0.0)$ \\
    \midrule
    \multicolumn{2}{l}{\textit{Stage~II (Align Training)}} \\
    \quad Trainable experts & video, latent dynamics, action \\
    \quad Views             & $3$ (head $+$ wrists) \\
    \quad Batch size / GPU  & $96$ \\
    \quad Schedule          & $1$ epoch \\
    \quad $(\lambda_v,\lambda_d,\lambda_a)$ & $(1.0,\,0.5,\,1.0)$ \\
    \midrule
    \multicolumn{2}{l}{\textit{Stage~III (Post-training)}} \\
    \quad Trainable experts & video, latent dynamics, action \\
    \quad Views             & target embodiment \\
    \quad Batch size / GPU  & $48$ \\
    \quad Schedule          & $5$ epochs \\
    \quad $(\lambda_v,\lambda_d,\lambda_a)$ & $(1.0,\,0.1,\,1.0)$ \\
    \bottomrule
  \end{tabular}
  }
  \caption{Training hyperparameters of the World Dynamics Action Model (WDAM).}
  \label{tab:train_hparams}
\end{table}

\paragraph{World Dynamics Action Model.}
Our WDAM is a three-expert Mixture-of-Transformers built on a
\textbf{Wan2.2 TI2V-5B} video-diffusion backbone. The video expert (a video DiT)
has $30$ layers, hidden dimension $3072$, $24$ attention heads of head dimension
$128$, and FFN dimension $14336$, operating on the $16$-channel latents of the
Wan2.2 causal VAE ($4\times$ temporal compression); the T5 text encoder and the
VAE are kept frozen, so only the video DiT is trained. Alongside it run two
lightweight experts fused into the video expert through per-layer mixed
self-attention ($1{:}1$ layer participation across all $30$ layers, with the
attention geometry, i.e., number of heads and head dimension, injected from the
video backbone): (i)~the \textbf{Latent Dynamics Expert}, a learnable-query
expert (residual width $512$, FFN $2048$) that emits $k_d{=}4$ latent-dynamics
tokens per predicted video latent ($2$ predicted latents $\Rightarrow 8$ tokens),
regressing the $512$-dimensional LDM target $z_i$; and (ii)~the
\textbf{Action Expert} (residual width $1024$, FFN $4096$) that flow-matches the
robot action, predicting $16$ actions per latent. The Action Expert shares a
single transformer body across heterogeneous embodiments---covering both
parallel-gripper and dexterous-hand robots---and adapts to each by swapping only
its input and output projection layers, so the shared backbone need not commit
to a fixed action dimension.
The latent-dynamics target $\{z_i\}$ is produced by the frozen LDM run on the raw
observation frames (online during Stage~I, precomputed offline thereafter); at
inference the LDM is not invoked and the Latent Dynamics Expert predicts $\{z_i\}$
directly.

The WDAM is trained under an inverse-dynamics protocol: video denoising is
teacher-forced during training, while at inference the action is solved from the
predicted video. The three experts are supervised jointly by
\begin{equation}
  \mathcal{L}_{\text{WDAM}} = \lambda_v\,\mathcal{L}_{\text{video}}^{\text{fm}}
              + \lambda_d\,\mathcal{L}_{\text{dyn}}^{\text{mse}}
              + \lambda_a\,\mathcal{L}_{\text{act}}^{\text{fm}}.
  \tag{\ref{eq:wdam_loss}}
\end{equation}

The three-stage curriculum shifts the weights $(\lambda_v,\lambda_d,\lambda_a)$
to reflect each stage's objective. \emph{Stage~I (Pre-training)} trains the video
and latent dynamics experts on single-view human and robot video with the action
expert frozen, $(\lambda_v,\lambda_d,\lambda_a){=}(1.0,\,1.0,\,0.0)$.
\emph{Stage~II (Align Training)} warm-starts from the Stage~I checkpoint and
unfreezes all experts, training on general robot actions over a $3$-view
(head $+$ two wrists) canvas, $(1.0,\,0.5,\,1.0)$ (the action expert and
proprioception encoder are (re)initialized rather than loaded). The three views
are composed into a single L-shape canvas---the head view spanning the full width
across the top ($2/3$ of the height) and the two wrist views placed side by side
along the bottom---so all cameras are modeled jointly within one video stream.
\emph{Stage~III (Post-training)} fine-tunes all experts on the target embodiment,
$(1.0,\,0.1,\,1.0)$.

\begin{figure}[t]
  \centering
  \includegraphics[width=\linewidth]{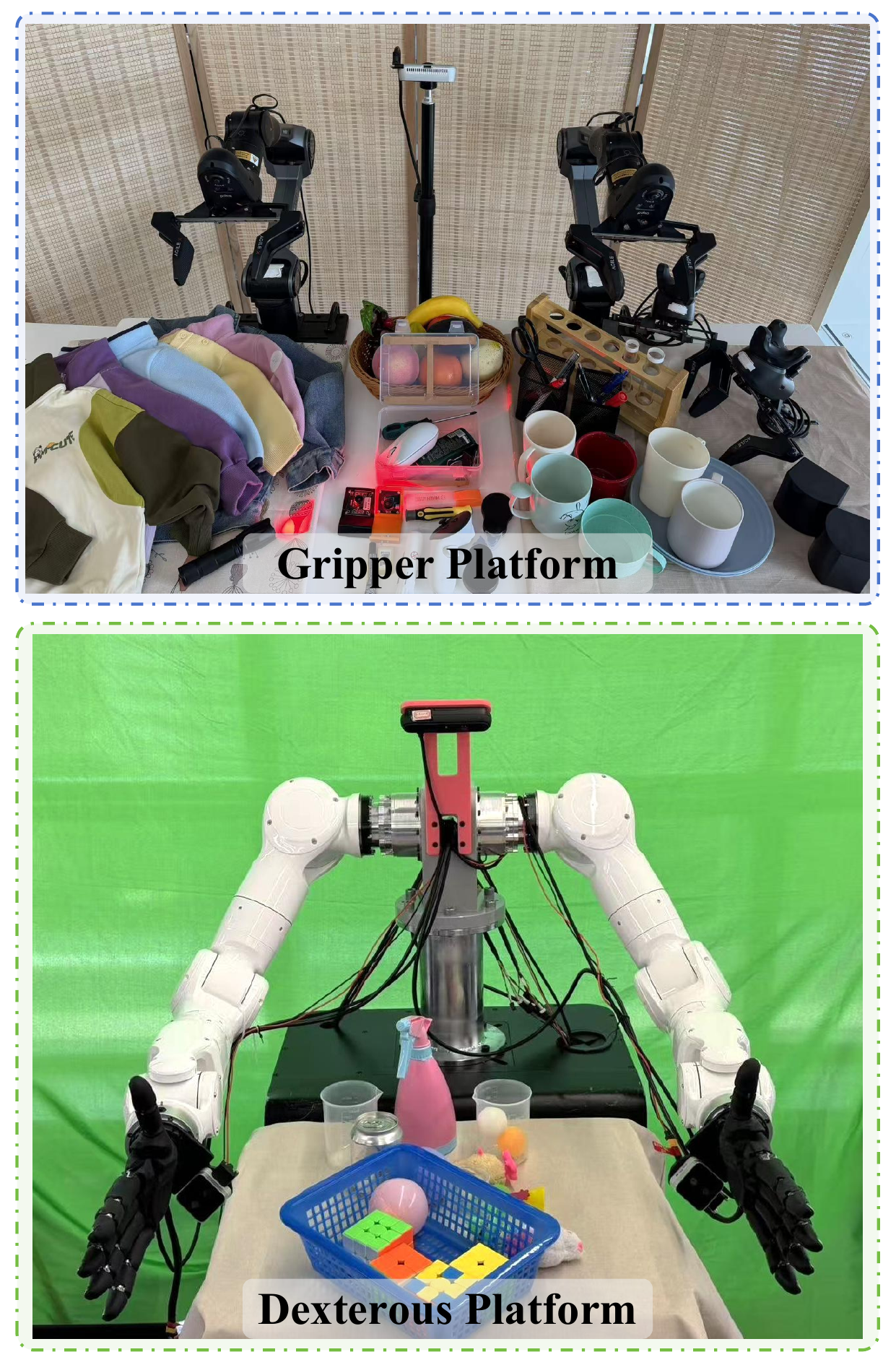}
  \caption{Real-robot platforms. Top: the gripper platform, a dual-arm AgileX PiPER
  with Pika Grippers. Bottom: the dexterous platform, dual Tianji arms with Wuji hands.}
  \label{fig:real_setup}
\end{figure}

All stages optimize with AdamW ($\beta_1{=}0.9,\ \beta_2{=}0.95$, weight decay
$0.01$) under a cosine learning-rate schedule with peak learning rate
$1\times10^{-4}$, $5\%$ linear warmup, and a minimum learning rate of $1\%$ of
the peak; gradients are clipped to a global norm of $1.0$. All training is conducted on $64$ NVIDIA H20 GPUs. We train in \texttt{bf16}
mixed precision with DeepSpeed ZeRO-2 sharding, gradient checkpointing, and a
fixed seed of $42$; the T5 text encoder, the Wan2.2 VAE, and
the LDM are frozen throughout. Stage-specific settings (per-GPU batch size,
schedule length, and loss weights) are summarized in Tab.~\ref{tab:train_hparams}.

\subsection{D. Real-World Setup and Task Description}

\paragraph{Hardware.}
We deploy on two real-robot platforms, shown in Fig.~\ref{fig:real_setup}.
The \emph{gripper platform} (top) is a dual-arm AgileX PiPER system, with an AgileX
Pika Gripper mounted on each arm. It observes the scene through three cameras: an
Intel RealSense D435i on the head and an Intel RealSense D405 on each wrist.
Demonstrations are teleoperated with two AgileX Pika Sense handles paired with two
AgileX Pika Stations.
The \emph{dexterous platform} (bottom) pairs two Tianji arms with two Wuji
multi-fingered dexterous hands, and uses an Orbbec Gemini~335L head camera together
with an Intel RealSense D405 on each wrist.

We evaluate on seven real-world manipulation tasks that span bimanual and
dexterous-hand settings and cover a range of skills, including precise pick-and-place,
fine-grained positional adjustment, articulated- and deformable-object manipulation,
cross-arm handover, dexterous multi-finger manipulation, and long-horizon multi-step
reasoning. Five are executed on the gripper platform and two on the dexterous
platform. Detailed instructions are given below.

\smallskip
\noindent\textbf{\large Gripper Platform}

\paragraph{Sorting.} \textit{Move the holder to the brown zone, put stationery inside,
then put fruit into the basket.} This is a long-horizon task that requires categorizing
multiple object types and placing them into their designated regions in the correct
order.

\paragraph{Shift Test Tube.} \textit{Shift the edge test tube inward by two slots.}
This task tests high-precision, small-scale positional control, as the target
displacement spans only two slots.

\paragraph{Tidy Desk.} \textit{Stow the clutter into the box and flip the lid closed.}
Beyond pick-and-place, the task requires articulated-object manipulation to close the
lid after all items are stowed.

\paragraph{Fold Shirt.} \textit{Fold the cloth.} This task involves deformable-object
manipulation, where the configuration of the cloth changes continuously throughout the
interaction.

\paragraph{Handover Mug.} \textit{Pick up the right cup, hand it over to the left arm,
and place it on the left plate.} This is a bimanual task that requires coordinated
cross-arm handover followed by accurate placement.

\smallskip
\noindent\textbf{\large Dexterous Platform}

\paragraph{Place Rubik's Cube.} \textit{From a clutter of diverse objects, locate the
Rubik's cube, pick it up, and place it into the basket.} This task couples target
identification among many distractors with a dexterous multi-finger grasp.

\paragraph{Spray Water.} \textit{The right hand picks up the spray bottle and passes it
to the left hand; the left hand reorients it and passes it back to the right hand,
which grips it and presses the trigger to spray water onto the plant.} This is a
bimanual dexterous task requiring in-hand tool use with independent finger actuation to
operate the trigger.

\subsection{E. Latent Dynamics Regression}
\label{app:ldm_reg}

We detail the regression probe used to evaluate the latent dynamics. Our goal is to
quantify how much action-relevant information the latent dynamics from our LDM already
encode. To this end, we keep the latent dynamics fixed and train only a shallow
two-layer MLP head to regress the motion from them. Because this head is deliberately shallow, its accuracy reflects the information already in the latent dynamics rather than capacity the head adds.

\paragraph{Data.}
We build the evaluation set from RoboTwin over all $50$ tasks, using $2{,}500$
clean and $2{,}500$ randomized episodes. RoboTwin is not part of the LDM training
corpus, so all variants are probed on an unseen domain. We sweep six sampling
rates: at stride $s$, each episode is cut into clips of eight frames taken $s$
apart, with consecutive clips offset by $8s$ frames so that they do not overlap.
Near-static transitions are discarded so that they do not dilute the error. The
remaining clips are split $8{:}2$ into probe-training and validation; the
resulting counts are listed in Tab.~\ref{tab:lam_full}, ranging from about
$0.68$M training clips at $50$\,Hz to $31$K at $2.5$\,Hz.

\begin{table}[t]
\centering
\caption{Latent-dynamics regression, full per-rate breakdown (scale-normalized MSE
$\downarrow$ on translation$+$rotation). Columns are sampling rates; best per column
in bold. Bottom rows list the train/validation clip counts per rate.}
\label{tab:lam_full}
\resizebox{\columnwidth}{!}{%
\begin{tabular}{@{}lcccccc@{}}
\toprule
 & \multicolumn{3}{c}{High-rate} & \multicolumn{3}{c}{Low-rate} \\
\cmidrule(lr){2-4}\cmidrule(l){5-7}
Method & 50\,Hz & 25\,Hz & 12.5\,Hz & 10\,Hz & 5\,Hz & 2.5\,Hz \\
\midrule
\textbf{LDM}            & \textbf{0.056} & \textbf{0.158} & \textbf{0.426} & \textbf{0.607} & \textbf{1.815} & \textbf{3.956} \\
\quad w/o motion align.   & 0.131          & 0.487          & 1.718          & 2.581          & 8.497          & 20.686 \\
\quad w/o multi-stride   & 0.057          & 0.170          & 0.482          & 0.712          & 2.205          & 4.700 \\
\quad w/ pixel recon.    & 0.063          & 0.188          & 0.503          & 0.742          & 2.260          & 5.070 \\
\quad w/ hard VQ         & 0.093          & 0.268          & 0.738          & 1.064          & 3.236          & 7.692 \\
\midrule
$n_{\text{train}}$ & 676{,}447 & 358{,}711 & 196{,}395 & 157{,}868 & 75{,}828 & 31{,}276 \\
$n_{\text{val}}$   & 169{,}413 & 89{,}808  & 49{,}070  & 39{,}489  & 18{,}928 & 7{,}763 \\
\bottomrule
\end{tabular}%
}
\vspace{-2mm}
\end{table}

\paragraph{Regression target.}
Each $8$-frame clip yields seven adjacent transitions. For each, the ground-truth
motion is the difference between consecutive absolute end-effector states: a
$12$-D per-step target stacking the translation ($\Delta p$) and rotation
($\Delta q$) of both arms. Each modality is divided by a fixed scale
($0.01\,\mathrm{m}$ for translation, $0.08\,\mathrm{rad}$ for rotation), matching the scale used for motion alignment during LDM training.

\paragraph{Head and optimization.}
Each model's latent is first mapped by a shared $\mathrm{Linear}(D\!\to\!512)$
projection (identity-initialized when $D=512$) that only removes cross-model
dimensionality differences, followed by a two-layer MLP
($\mathrm{LN}\!\to\!512\!\to\!256\!\to\!12$, GELU, dropout). All models use this same
head and the same optimization: AdamW with learning rate $10^{-3}$, weight decay
$10^{-4}$, batch size $1024$, and $200$ epochs under a cosine schedule with no early
stopping. A separate head is trained per sampling rate.

\begin{figure*}[!t]
    \centering
    \includegraphics[width=\textwidth]{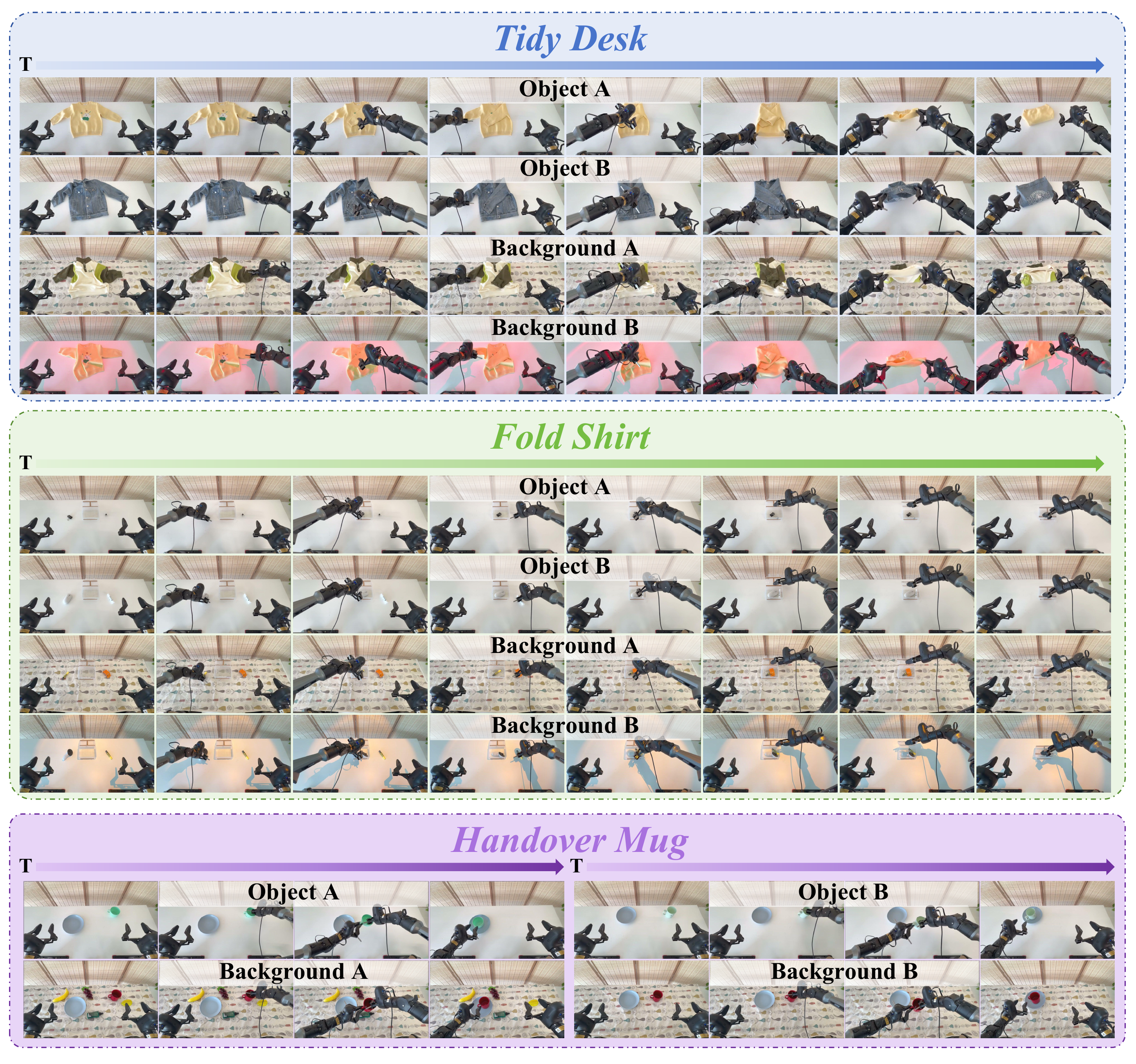}
    \caption{Real-world Generalization Rollouts. Qualitative rollouts of our policy on three real-world tasks, Tidy Desk,
Fold Shirt, and Handover Mug. Time proceeds from left to right
along the arrow ($\mathbf{T}$). For each task we test robustness under object
generalization (Object~A / Object~B, unseen target instances) and
background generalization (Background~A / Background~B, unseen
tabletop textures, clutter, and lighting).}
    \label{fig:realdeployment_more}
\end{figure*}

\paragraph{Rate bands.}
RoboTwin renders at a native $50$\,Hz, and each rate is obtained by taking frames
at a fixed stride from this source. The six rates are grouped into a \emph{high-rate}
band ($50/25/12.5$\,Hz, stride $1/2/4$) and a \emph{low-rate} band ($10/5/2.5$\,Hz,
stride $5/10/20$), which probe the latent over small and large inter-frame motions,
respectively. The full per-rate breakdown is given in Tab.~\ref{tab:lam_full}.

\subsection{F. RoboTwin Detailed Results}

Tab.~\ref{tab:robotwin} reports the per-task success rates on all 50 RoboTwin
tasks under both the clean and randomized evaluation settings, complementing the
aggregate numbers in the main paper. Our method attains the highest average
success rate under both settings (\textbf{93.96} clean / \textbf{92.78}
randomized).

\subsection{G. More Real-world Deployment Visualizations}

Fig.~\ref{fig:realdeployment_more} shows additional real-world rollouts of our policy on three tasks, \emph{Tidy Desk}, \emph{Fold Shirt}, and
\emph{Handover Mug}. For each task we test generalization to unseen objects and unseen backgrounds. The policy completes every
task under all variations, showing robustness to novel objects and scenes.

\begin{table*}[t]
\centering
\caption{Per-task success rates on RoboTwin under clean and randomized evaluation settings.}
\label{tab:robotwin}
\footnotesize
\setlength{\tabcolsep}{2.5pt}
\begin{tabular}{l cc cc cc cc cc cc cc}
\toprule
\multirow{2}{*}{Task} & \multicolumn{2}{c}{\textbf{Ours}} & \multicolumn{2}{c}{Fast-WAM} & \multicolumn{2}{c}{Lingbot-VA} & \multicolumn{2}{c}{$\pi_{0.5}$} & \multicolumn{2}{c}{Motus} & \multicolumn{2}{c}{LaWAM} & \multicolumn{2}{c}{ACE-Ego-0} \\
\cmidrule(lr){2-3}\cmidrule(lr){4-5}\cmidrule(lr){6-7}\cmidrule(lr){8-9}\cmidrule(lr){10-11}\cmidrule(lr){12-13}\cmidrule(lr){14-15}
 & Clean & Rand. & Clean & Rand. & Clean & Rand. & Clean & Rand. & Clean & Rand. & Clean & Rand. & Clean & Rand. \\
\midrule
Adjust Bottle              & \textbf{100} & \textbf{100} & \textbf{100} & \textbf{100} & 90 & 94 & \textbf{100} & 99 & 89 & 93 & \textbf{100} & \textbf{100} & \textbf{100} & \textbf{100} \\
Beat Block Hammer          & 95 & 97 & \textbf{99} & 97 & 96 & \textbf{98} & 96 & 93 & 95 & 88 & 90 & 93 & 98 & 92 \\
Blocks Ranking RGB         & \textbf{100} & \textbf{100} & \textbf{100} & \textbf{100} & 99 & 98 & 92 & 85 & 99 & 97 & 97 & \textbf{100} & 98 & 97 \\
Blocks Ranking Size        & 79 & 69 & \textbf{94} & \textbf{98} & \textbf{94} & 96 & 49 & 26 & 75 & 63 & 93 & 89 & 89 & 91 \\
Click Alarmclock           & 98 & 99 & \textbf{100} & \textbf{100} & 99 & \textbf{100} & 98 & 89 & \textbf{100} & \textbf{100} & \textbf{100} & \textbf{100} & 52 & 38 \\
Click Bell                 & \textbf{100} & \textbf{100} & \textbf{100} & \textbf{100} & \textbf{100} & \textbf{100} & 99 & 66 & \textbf{100} & \textbf{100} & \textbf{100} & \textbf{100} & 66 & 71 \\
Dump Bin Bigbin            & 99 & \textbf{97} & 97 & 96 & 89 & 96 & 92 & \textbf{97} & 95 & 91 & 97 & 95 & \textbf{100} & \textbf{97} \\
Grab Roller                & \textbf{100} & 99 & \textbf{100} & \textbf{100} & \textbf{100} & \textbf{100} & \textbf{100} & \textbf{100} & \textbf{100} & \textbf{100} & \textbf{100} & \textbf{100} & \textbf{100} & \textbf{100} \\
Handover Block             & \textbf{100} & \textbf{89} & 95 & 81 & 99 & 78 & 66 & 57 & 86 & 73 & 96 & 87 & 96 & 85 \\
Handover Mic               & \textbf{99} & 97 & \textbf{99} & \textbf{100} & 94 & 96 & 98 & 97 & 78 & 63 & 93 & 98 & 91 & 94 \\
Hanging Mug                & 37 & 49 & \textbf{58} & \textbf{62} & 40 & 28 & 18 & 17 & 38 & 38 & 51 & 43 & 29 & 31 \\
Lift Pot                   & \textbf{100} & \textbf{100} & \textbf{100} & \textbf{100} & \textbf{100} & 99 & 96 & 85 & 96 & 99 & \textbf{100} & 99 & \textbf{100} & \textbf{100} \\
Move Can Pot               & 99 & 97 & 90 & 88 & 94 & 97 & 51 & 55 & 34 & 74 & 98 & 93 & \textbf{100} & \textbf{98} \\
Move Pillbottle Pad        & 99 & \textbf{100} & \textbf{100} & 99 & 99 & 99 & 84 & 61 & 93 & 96 & 97 & 90 & \textbf{100} & \textbf{100} \\
Move Playingcard Away      & \textbf{100} & \textbf{100} & \textbf{100} & \textbf{100} & \textbf{100} & 99 & 96 & 84 & \textbf{100} & 96 & \textbf{100} & \textbf{100} & \textbf{100} & 98 \\
Move Stapler Pad           & \textbf{96} & \textbf{89} & 77 & 64 & 91 & 79 & 56 & 42 & 83 & 85 & 94 & 87 & 90 & \textbf{89} \\
Open Laptop                & 99 & \textbf{100} & 98 & \textbf{100} & 92 & 94 & 90 & 96 & 95 & 91 & \textbf{100} & \textbf{100} & \textbf{100} & 98 \\
Open Microwave             & 79 & 73 & 62 & 45 & 82 & 86 & 34 & 77 & \textbf{95} & \textbf{91} & 41 & 43 & 91 & 85 \\
Pick Diverse Bottles       & 84 & 89 & 80 & 85 & 89 & 82 & 81 & 71 & 90 & \textbf{91} & \textbf{91} & 88 & 84 & 86 \\
Pick Dual Bottles          & 99 & \textbf{100} & \textbf{100} & 96 & \textbf{100} & 99 & 93 & 63 & 96 & 90 & \textbf{100} & 95 & 89 & 88 \\
Place A2B Left             & \textbf{99} & 95 & 95 & 93 & 97 & 93 & 87 & 82 & 88 & 79 & 98 & 91 & 95 & \textbf{96} \\
Place A2B Right            & \textbf{99} & 95 & 93 & \textbf{99} & 97 & 95 & 87 & 84 & 91 & 87 & 89 & 94 & 90 & 94 \\
Place Bread Basket         & 88 & \textbf{99} & 91 & 93 & \textbf{97} & 95 & 77 & 64 & 91 & 94 & 92 & 85 & 92 & 93 \\
Place Bread Skillet        & \textbf{99} & \textbf{96} & 90 & 93 & 95 & 90 & 85 & 66 & 86 & 83 & 90 & 83 & 94 & 89 \\
Place Burger Fries         & \textbf{99} & 99 & 96 & 99 & 97 & 95 & 94 & 87 & 98 & 98 & 93 & 96 & 98 & \textbf{100} \\
Place Can Basket           & 85 & 80 & 71 & 69 & 81 & \textbf{84} & 62 & 62 & 81 & 76 & \textbf{92} & 65 & 78 & 82 \\
Place Cans Plasticbox      & 95 & 97 & 99 & 96 & \textbf{100} & \textbf{99} & 94 & 84 & 98 & 94 & \textbf{100} & 95 & \textbf{100} & 98 \\
Place Container Plate      & 99 & 99 & 96 & \textbf{100} & 99 & 97 & 99 & 95 & 98 & 99 & \textbf{100} & \textbf{100} & 98 & \textbf{100} \\
Place Dual Shoes           & 91 & \textbf{96} & 94 & 88 & 94 & 89 & 75 & 75 & 93 & 87 & \textbf{98} & 94 & 95 & \textbf{96} \\
Place Empty Cup            & \textbf{100} & \textbf{100} & \textbf{100} & \textbf{100} & \textbf{100} & \textbf{100} & \textbf{100} & 99 & 99 & 98 & 99 & \textbf{100} & \textbf{100} & \textbf{100} \\
Place Fan                  & 95 & \textbf{97} & 96 & 96 & \textbf{99} & 93 & 87 & 85 & 91 & 87 & 92 & 93 & 94 & 93 \\
Place Mouse Pad            & \textbf{98} & \textbf{99} & 83 & 89 & 93 & 96 & 60 & 39 & 66 & 68 & 91 & 84 & 96 & 95 \\
Place Object Basket        & \textbf{97} & \textbf{97} & 89 & 88 & 91 & 88 & 80 & 76 & 81 & 87 & 92 & 90 & 93 & 89 \\
Place Object Scale         & \textbf{100} & 91 & 90 & \textbf{97} & 96 & 95 & 86 & 80 & 88 & 85 & 95 & 88 & 95 & 92 \\
Place Object Stand         & 89 & \textbf{97} & 90 & 94 & \textbf{99} & 96 & 91 & 85 & 98 & \textbf{97} & 92 & 93 & 95 & 94 \\
Place Phone Stand          & 93 & 98 & \textbf{97} & \textbf{99} & \textbf{97} & 97 & 81 & 81 & 87 & 86 & 93 & 94 & 91 & 98 \\
Place Shoe                 & 99 & \textbf{100} & 96 & 99 & 98 & 98 & 92 & 93 & 99 & 97 & \textbf{100} & \textbf{100} & \textbf{100} & \textbf{100} \\
Press Stapler              & 97 & 85 & 90 & 97 & 85 & 82 & 87 & 83 & 93 & \textbf{98} & \textbf{98} & 97 & \textbf{98} & \textbf{98} \\
Put Bottles Dustbin        & 92 & \textbf{97} & \textbf{95} & 90 & 87 & 91 & 84 & 79 & 81 & 79 & 94 & 92 & 94 & 93 \\
Put Object Cabinet         & 77 & 80 & \textbf{94} & \textbf{89} & 85 & 87 & 80 & 79 & 88 & 71 & 90 & 82 & 82 & 79 \\
Rotate QRcode              & \textbf{99} & 91 & 93 & 89 & 96 & 91 & 89 & 87 & 89 & 73 & 94 & 89 & 94 & \textbf{95} \\
Scan Object                & 95 & 95 & 89 & 92 & \textbf{96} & 91 & 72 & 65 & 67 & 66 & \textbf{96} & 90 & 95 & \textbf{97} \\
Shake Bottle               & \textbf{100} & \textbf{100} & \textbf{100} & \textbf{100} & \textbf{100} & 97 & 99 & 97 & \textbf{100} & 97 & \textbf{100} & \textbf{100} & \textbf{100} & \textbf{100} \\
Shake Bottle Horizontally  & \textbf{100} & \textbf{100} & \textbf{100} & \textbf{100} & \textbf{100} & 99 & 99 & 99 & \textbf{100} & 98 & \textbf{100} & \textbf{100} & \textbf{100} & \textbf{100} \\
Stack Blocks Three         & 97 & 93 & 95 & 97 & \textbf{99} & \textbf{98} & 91 & 76 & 91 & 95 & 90 & 75 & 87 & 82 \\
Stack Blocks Two           & \textbf{100} & \textbf{100} & \textbf{100} & \textbf{100} & \textbf{100} & 98 & 97 & \textbf{100} & \textbf{100} & 98 & \textbf{100} & 97 & \textbf{100} & \textbf{100} \\
Stack Bowls Three          & 87 & 80 & 80 & 81 & 86 & 83 & 77 & 71 & 79 & \textbf{87} & \textbf{90} & 80 & 80 & 85 \\
Stack Bowls Two            & 99 & 95 & 92 & 98 & 94 & 98 & 95 & 96 & 98 & 98 & \textbf{100} & \textbf{99} & 96 & 98 \\
Stamp Seal                 & \textbf{97} & 99 & 90 & 94 & 96 & 97 & 79 & 55 & 93 & 92 & 89 & 88 & 94 & \textbf{100} \\
Turn Switch                & 71 & 45 & 61 & 59 & 44 & 45 & 62 & 54 & \textbf{84} & \textbf{78} & 47 & 56 & 59 & 57 \\
\midrule
\textbf{Average}           & \textbf{93.96} & \textbf{92.78} & 91.88 & 91.78 & 92.90 & 91.50 & 82.74 & 76.76 & 88.66 & 87.02 & 92.64 & 89.80 & 91.12 & 90.62 \\
\bottomrule
\end{tabular}
\end{table*}